\documentclass[10pt,twocolumn,letterpaper]{article}

\usepackage[pagenumbers]{cvpr} 

\usepackage[dvipsnames]{xcolor}

\definecolor{cvprblue}{rgb}{0.21,0.49,0.74}
\usepackage[pagebackref,breaklinks,colorlinks,citecolor=cvprblue]{hyperref}
\usepackage{booktabs}
\usepackage{multirow}
\usepackage{makecell}
\usepackage{multirow}
\usepackage{makecell}
\usepackage{graphicx}
\usepackage[dvipsnames]{xcolor}         

\def\paperID{11043} 
\def\confName{CVPR}
\def\confYear{2025}

\title{A Simple yet Effective Layout Token in Large Language Models for Document Understanding}

\author{
Zhaoqing Zhu$^{1*}$,
Chuwei Luo$^{1*\dagger}$, 
Zirui Shao$^{2*}$, 
Feiyu Gao$^{1\dagger}$, 
Hangdi Xing$^2$, 
Qi Zheng$^1$ \\
Ji Zhang$^1$ \\
$^1$Alibaba Group, 
$^2$Zhejiang University\\
{\tt\small \{zzhaoqing.z,luochuwei,zhengqisjtu\}@gmail.com}\\
\small{\texttt{\{shaozirui,xinghd\}@zju.edu.cn}}, \small{\texttt{\{feiyu.gfy,zj122146\}@alibaba-inc.com}}
}

\begin{document}
\maketitle
\def\thefootnote{*}\footnotetext{Equal contribution.$\dagger$ Corresponding author.}
\def\thefootnote{\arabic{footnote}}

\begin{abstract}
Recent methods that integrate spatial layouts with text for document understanding in large language models (LLMs) have shown promising results. A commonly used method is to represent layout information as text tokens and interleave them with text content as inputs to the LLMs. However, such a method still demonstrates limitations, as it requires additional position IDs for tokens that are used to represent layout information. Due to the constraint on max position IDs, assigning them to layout information reduces those available for text content, reducing the capacity for the model to learn from the text during training, while also introducing a large number of potentially untrained position IDs during long-context inference, which can hinder performance on document understanding tasks.
To address these issues, we propose LayTokenLLM, a simple yet effective method for document understanding. LayTokenLLM represents layout information as a single token per text segment and uses a specialized positional encoding scheme. It shares position IDs between text and layout tokens, eliminating the need for additional position IDs. This design maintains the model's capacity to learn from text while mitigating long-context issues during inference. Furthermore, a novel pre-training objective called Next Interleaved Text and Layout Token Prediction (NTLP) is devised to enhance cross-modality learning between text and layout tokens. Extensive experiments show that LayTokenLLM outperforms existing layout-integrated LLMs and MLLMs of similar scales on multi-page document understanding tasks, as well as most single-page tasks. 

\end{abstract}

\section{Introduction}
\label{sec:intro}

\begin{figure}[tp]
\centering
    \includegraphics[width=0.8\linewidth]{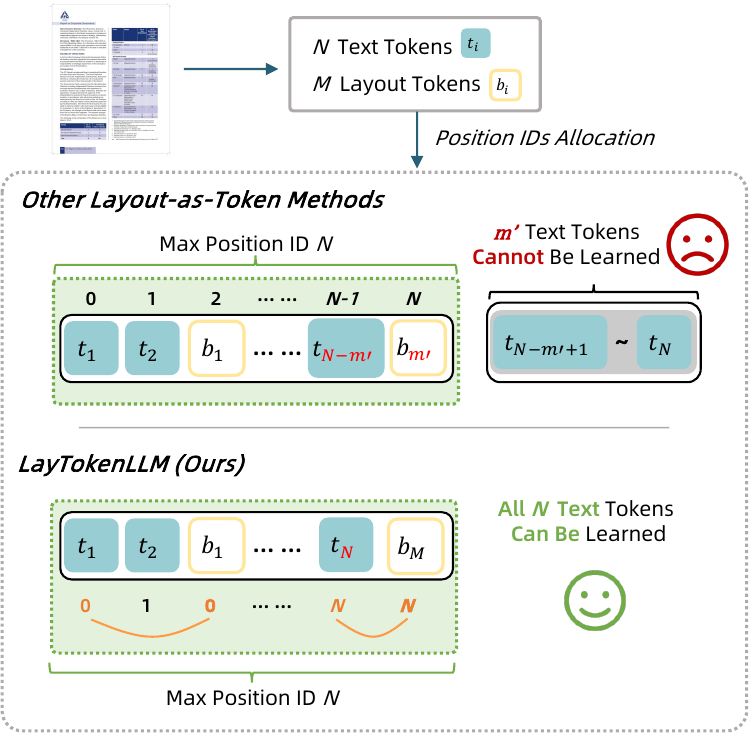}
    \vspace{-2mm}
    \caption{
      Comparison with other Layout-as-Token methods. 
       Previous Layout-as-Token methods require additional position IDs for layout information which squeeze the learning space for text content, 
       while LayTokenLLM eliminates the need for additional position IDs of layout information by sharing the first position ID of corresponding text content.
        }
    \label{motivation_figure}
    \vspace{-7mm}
  \end{figure}

  \begin{table*}[h!]
    \centering
    \small
  \begin{tabular}{lccc}
      
      \hline
      \textbf{Text and Layout Format} &
        \multicolumn{2}{l}{\textbf{\begin{tabular}[c]{@{}l@{}}Number of Extra Position IDs\\for Representing Layout\end{tabular}}} &
        \multicolumn{1}{l}{\begin{tabular}[c]{@{}c@{}}\textbf{T-Ratio}\\ ($N_{t}$ / N)\end{tabular}} \\ 
    \cline{2-3}
    & \small{In a Segment} & \small{Avg. on MP-DocVQA} &  \\ 
      \hline
      Plain Text (w/o Layout) & 0 & 0 & 100\% \\
      \hline
    \scalebox{0.92}{\{text:``text",Box:[123, 456, 133, 500]\}}~\cite{he2023icl}          & 27 & 8015 & 27.02\%           \\
    \scalebox{0.92}{$<$ref$>$text$<$/ref$><$box$>$(123,456),(133,500)$<$/box$>$}~\cite{bai2023qwenvl,liu2024textmonkey}    & 21 & 7959 & 32.26\%           \\
    \scalebox{0.92}{$<$ref$>$text$<$/ref$>$$<$box$>$[[253, 231, 733, 787]]$<$/box$>$}~\cite{chen2023internvl} & 18 & 6894 & 35.71\% \\ 
    
    text+one box hidden representation~\cite{lu2024boundingboxworthtoken}              & 1 & 350 & 90.91\% \\ \hline
    text+layout\_token (\textbf{Ours})                       & \textbf{0}  & \textbf{0}&  \textbf{100\%} \\ \hline
    \end{tabular}
    \vspace{-1mm}
    \caption{Comparison of different paradigm to integrate layout information with text content. 
    T-Ratio is defined as the ratio of the position utilization for text tokens ($N_t$) to the maximum trained position length ($N$). In the table, $N$ is set to 2048.
    }
    \label{table:comparison}
    \vspace{-4mm}
  \end{table*}

Document understanding~\cite{cui2021document} is currently an important area in both industry and academic research, driven by the critical need to efficiently process and understand complex documents.
In recent years, large language models (LLMs) and multimodal large language models (MLLMs) have made remarkable progress in this field.
Especially in tasks involving rich textual content, such as the document-oriented Visual Question Answering (DocVQA)~\cite{mathew2021docvqa} task and the visually-rich document information extraction (VIE)~\cite{jaume2019funsd,park2019cord,yang2023modeling} task.
Some works \cite{docllm,lapdoc,lu2024boundingboxworthtoken} suggest that the most valuable information for document understanding can be derived from both the text and its layout, treating spatial layouts as a form of lightweight visual information.
Building on this idea, these approaches \cite{docllm,lapdoc,lu2024boundingboxworthtoken} that integrate such spatial layouts as visual information with text for LLMs have shown promising results, and sometimes superior performance compared to MLLMs.

The integration of layout information can be broadly categorized into two types: layout-as-modality methods~\cite{docllm,luo2024layoutllm} and layout-as-tokens methods~\cite{he2023icl,perot2024lmdx,lu2024boundingboxworthtoken,lapdoc}.
Layout-as-modality methods treat layout information as an additional modality~\cite{docllm,luo2024layoutllm}, modeling it alongside text within LLMs and training specialized LLMs. Although layout-as-modality methods have shown good performance, they require substantial modifications to the LLM architecture to incorporate layout information, making them less lightweight. 
On the other hand, layout-as-tokens methods represent the layout information as text tokens, and interleave them with the corresponding text content as inputs into the LLMs, which provide a more lightweight and commonly used approach~\cite{lapdoc,lu2024boundingboxworthtoken} for document understanding.

However, existing layout-as-token methods still encounter a significant limitation.
Due to the constraint on max position IDs, assigning them to layout information reduces those available for text content, reducing the capacity for the model to learn from the text during training. 
As illustrated in \cref{motivation_figure}, the context window during training is constrained by the maximum position ID $N$.
When tokens that are used for representing layout information are integrated into an LLM, the allocation of additional position IDs ($m'$) for layout information reduces the number of position IDs available for the text content ($N-m'$), leading to less capacity for LLMs to learn from the text during training.
To better quantify the impact of incorporating additional layout information, a rough measure called T-Ratio which is used to represent the ratio of the position utilization for text tokens ($N_t$) to the maximum trained position length ($N$) is shown in \cref{table:comparison}.
As can be seen, assigning position IDs to tokens that are used for representing layout information significantly affects the T-ratio, even when only a single position ID is used to represent the layout in a segment.
Although the T-ratio is a rough measure, it reflects the impact of introducing layout information, as assigning position IDs to layout tokens reduces the number of position IDs available for text content, ultimately limiting the model's capacity to learn from the text during training.
And it can be also seen that existing methods allocate hundreds or even thousands of additional position IDs for layout information on the MP-DocVQA dataset and these additional position IDs are potentially unlearned during training, which may exacerbate the long-context inference problem~\cite{song2024hierarchical,peng2023yarn}.

To address these issues, we propose a simple yet effective framework in this paper, called LayTokenLLM, for document understanding.
LayTokenLLM is a lightweight framework that represents the layout information of each text segment as a single layout token and employs a specially designed positional encoding scheme for the layout tokens.
Notably, as shown in \cref{table:comparison}, LayTokenLLM incorporates layout information as a single layout token but without allocating any additional position ID, ensuring comprehensive learning of text content (100\% max position IDs for text content) during training, while alleviating long-context issues introduced by layout information during inference.
Additionally, a novel pre-training objective called Next Interleaved Text and Layout Token Prediction (NTLP) is proposed to improve the comprehension of interleaved format and deepen the connection between these distinct types of information in LayTokenLLM. Different from previous methods that focus solely on either text or layout content for subsequent predictions~\cite{lu2024boundingboxworthtoken,docllm}, NTLP leverages the auto-regressive traits of LLMs and additionally facilitates cross-prediction between text and layout.
Extensive experiments across widely used benchmarks for both single-page and multi-page document understanding demonstrate the effectiveness of the proposed LayTokenLLM.

Our contributions are summarized as follows:
\begin{itemize} \setlength{\itemsep}{0pt}
    \item[1)] This paper introduces LayTokenLLM, a simple yet effective method to integrate layout information into LLMs for document understanding. It represents layout information as a single token and uses a specially designed positional encoding scheme, avoiding the issues caused by allocating additional position IDs for the layout information.
    \item[2)] A novel pre-training objective called Next Interleaved Text and Layout Token Prediction is introduced to enhance cross-modal prediction and relational learning between text and layout modalities.
    \item[3)] Experimental results show that the proposed LayTokenLLM significantly outperforms existing methods utilizing LLMs/MLLMs for multi-page document understanding, while also achieving superior performance in most sub-tasks of single-page document comprehension. 
\end{itemize}

\section{Related Work}
Recently, leveraging large language models (LLMs) and multimodal large language models (MLLMs) 
 for document understanding have shown significant progress.
Although existing MLLMs show promising results in document understanding, they still struggle with issues associated with high-resolution input, particularly in cases of dense or difficult-to-recognize text.
Considering the layout information is vital for document understanding~\cite{xu2020layoutlm,xu2020layoutlmv2,huang2022layoutlmv3,li2021structurallm,bivldoc,li2021structext,appalaraju2021docformer,gu2022xylayoutlm,yu2023structextv,Luo_2023_CVPR,da2023vision,proctag},
existing an alternative approach, integrating spatial layouts with text as lightweight visual information for LLMs has shown promising results, and sometimes even superior performance compared to MLLMs.
These approaches can be categorized into two types: layout-as-modality methods and layout-as-tokens methods.
\subsection{Multi-modal Large Language Models}
Existing MLLMs \citep{gpt4v,team2023gemini,NEURIPS2023_6dcf277e,ye2024mplug,qwen-vl,internvl,zhu2024minigpt} show exceptional performance for document understanding.
Models \citep{qwen-vl,internvl,chen2024far,zhang2024long,li2024llava} like InternVL, Qwen-VL have augmented MLLMs with advanced visual capabilities by introducing high-resolution visual input to better handle documents containing dense or difficult-to-recognize text.
However, the methods require an excessive number of image tokens, adversely affecting inference speed~\cite{chen2024far,zhang2024long,li2024llava}. 
In response to this challenge, a series of MLLMs \citep{hu2024mplug,li2024tokenpacker,hu2024mplug1.5} propose to reduce the token count by compressing image patches, but this may lead to the loss of critical textual information.

\subsection{Layout-as-Modality Methods}
Layout-as-Modality methods treat layout information as an additional modality, modeling it alongside text within LLMs \citep{gpt3, llama, qwen1.5} and training specialized LLMs for document understanding. \citet{luo2024layoutllm} make pioneering attempts to combine layout information with text into LLMs. In order to fully exploit the document layout information, it employs pre-trained document encoders, which represent the spatial layout of text as an additional modality similar to previous pre-trained text-layout models \cite{Xu2020LayoutLMv2MP, layoutlmv3}. Recently, \citet{docllm} propose to further disentangle the text and layout modalities by considering the inter-dependency between them. However, these methods require modifying the model architecture and necessitate an additional pre-training stage, making them less lightweight.

\subsection{Layout-as-Token Methods}
Layout-as-Tokens methods represent layout information as text sequences, embedding the sequences interleaved with the corresponding text as inputs into the LLMs as shown in \cref{table:comparison}, providing a more natural and commonly used approach. 
Specifically, \citet{he2023icl} introduce an in-context learning format like ``\textit{\{text:``text",Box:[123, 456, 133, 500]\}}" which incorporate layout information (See \cref{table:comparison}, line 2) in the demonstration to enable LLMs to understand positional relationships. 
And \citet{lapdoc} design a novel document verbalizer to effectively encode the layout information in the prompt. \citet{perot2024lmdx} generate LLM prompts containing both the text
content and coordinate tokens, which communicate the layout modality and act as unique identifiers of the text segments for information extraction and localization, while \citet{lu2024boundingboxworthtoken} use one hidden token to represent layout information.
Despite their convenience and effectiveness, these methods introduce an excessive number of interleaved position spaces to represent the layout, leading to a dilution of the textual content (see \cref{table:comparison}). 
The extra interleaved position occupied by models not only hampers the comprehension learning and increases the burden of comprehension of the text content.

\section{Method}
In this section, our LayTokenLLM is presented, which is an LLM-based method that incorporates text and spatial layout information which can be viewed as lightweight visual information. 
To incorporate layout information while avoiding issues arising from extra position ID allocations, and to enhance the connection between text and layout within the same segment, two primary components are proposed: a simple yet effective Layout Token, and a pre-training objective designed for interleaved text and layout format.
\begin{figure*}[tp]
    \centering
    \includegraphics[width=0.83\linewidth]{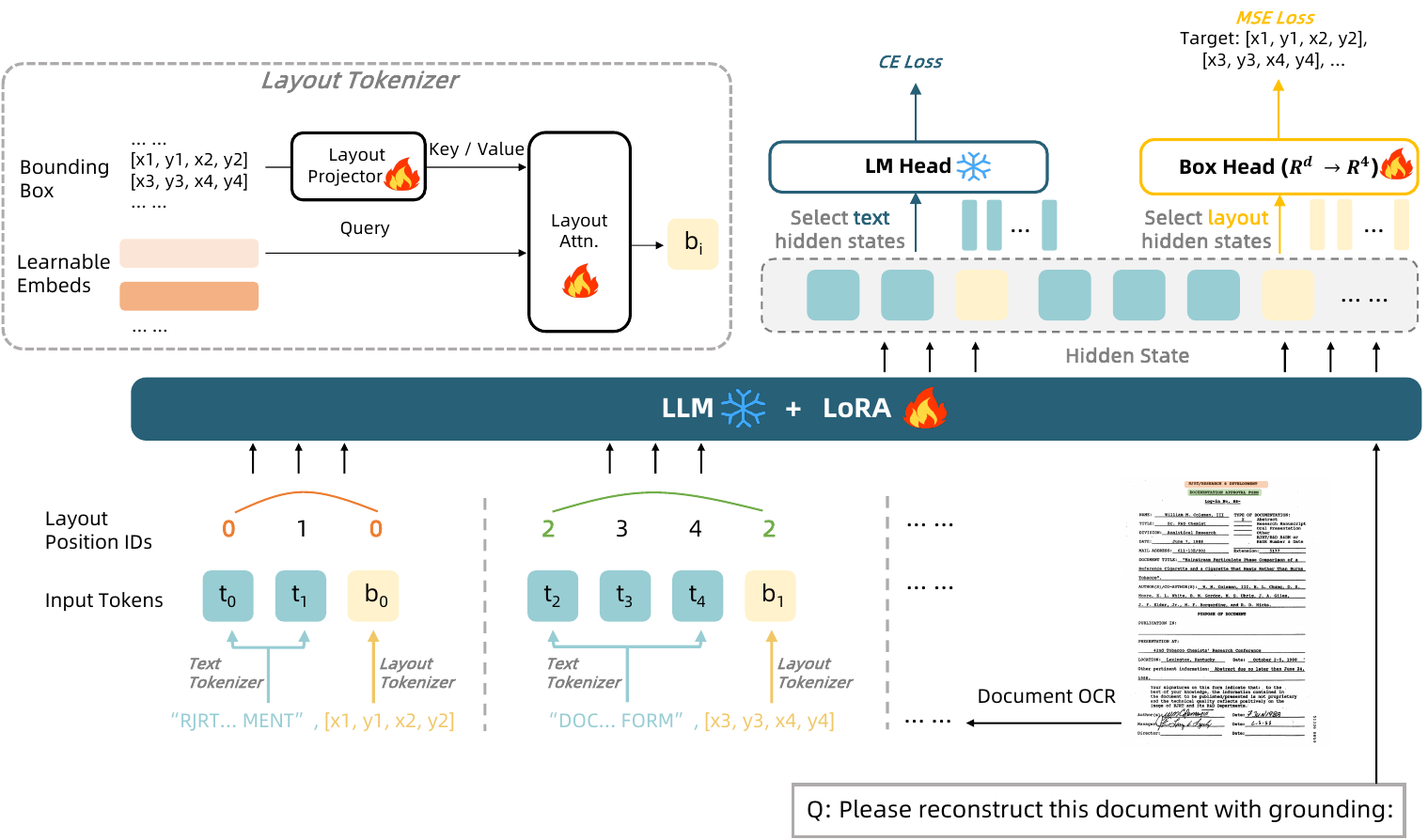}
    \vspace{-1mm}
    \caption{
      The overall architecture of LayTokenLLM. 
      Given the text segments with layouts parsed from document (e.g., by OCR), LayTokenLLM first tokenizes the layout information (bounding box) of each text segment into a single layout token by leveraging a trainable projector and an attention module with learnable query.
      Subsequently, the text tokens and layout tokens are interleaved and the position IDs are assigned by sharing the first position ID of each text segment with the corresponding layout token, preserving the entire learning space for textual content. Finally, distinct training objectives are employed for the text and layout information, respectively.
        }
    \label{pipeline_figure}
    \vspace{-4mm}
\end{figure*}

\subsection{Model Architecture}
The overall architecture of LayTokenLLM is shown in \cref{pipeline_figure}.
Once the text segments with corresponding layout information are parsed from the document (e.g., by OCR), the bounding box coordinates of each text segment are first compressed into one single layout token with a layout tokenizer. Then the text tokens and their corresponding layout tokens are interleaved and input to LLM. 
A simple yet effective layout positional encoding scheme is designed to address the issues of additional position IDs. Furthermore, a novel pretraining objective is proposed to enhance cross-modal connections within the same segment.

\subsubsection{Details of Layout Token}
As shown in the upper left part of \cref{pipeline_figure}, a learnable embedding \(t \in \mathbb{R}^{d}\) is employed as a query, mapping each text segment's bounding box \(Box\) into only one single layout token \(b \in \mathbb{R}^{d}\):
\begin{equation}
b = F_{Attn}(t, F_B(Box)),
\label{eq:lay_token}
\end{equation}
where \(F_B\) represents a projector that encodes the bounding box defined by four-dimensional coordinates \([x1, y1, x2, y2]\) into a high-dimensional embedding, and \(F_{Attn}\) represents an attention encoder which takes the learnable embedding as query and the high-dimensional embedding of bounding box as key and value.
Through the layout tokenizer, the layout information is  significantly compressed, thereby alleviating the burden of longer tokens while enhancing inference speed.

\subsubsection{Positional Encoding Scheme for Layout Token}
The most prevalent positional encoding method for LLMs is the Rotary Positional Encoding (RoPE)~\cite{su2024roformer}. 
Let \(T\) and \(L\) represent the length of tokens used for text and layout information in an OCR segment, previous methods will allocate additional position IDs for the interleaved layout information and set the position IDs of a segment \(P\) as:
\begin{equation}
P = [0, 1, \ldots, T-1, T, \ldots, T+L-1].
\label{eq:ori_rope_pos_id}
\end{equation}
However, even compressing the layout information to a single layout token for each text segment, an additional position ID must still be allocated. Moreover, the positional distance between adjacent text segments will be stretched due to the inserted layout tokens.

To address the issues of additional position IDs and the comprehension burden of stretched positional distance introduced by layout information, a straightforward and efficient positional encoding scheme is proposed that reuses the position IDs already utilized in the text tokens for layout tokens.
Considering the cross-modality alignment within the same text segment, each single layout token is assigned with the position ID of the first text token in its corresponding text content (as illustrated in the lower left part of \cref{pipeline_figure}). Then the position IDs of a text segment \(P\) is expressed as:
\begin{equation}
P = [\textcolor{red}{0}, 1, \ldots, T-1, \textcolor{red}{0}].
\label{eq:our_pos_id}
\end{equation}

Consequently, LayTokenLLM needs no additional position IDs for layout information, enabling the trained position IDs to be entirely dedicated to text content, and achieving a 100\% T-Ratio. At the same time, the positional distance between adjacent text segments is preserved.

\subsection{Pretraining Objective}
\begin{figure}[tp]
\centering
  \includegraphics[width=0.9\linewidth]{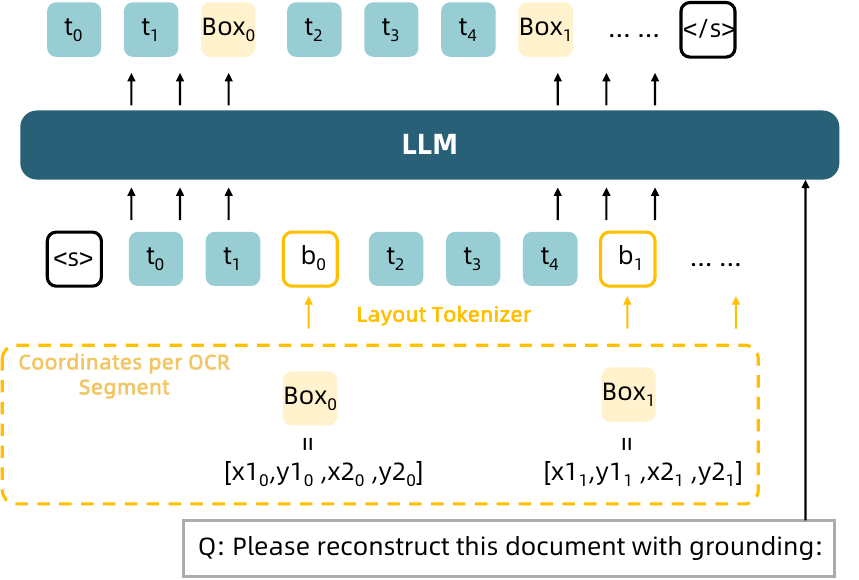}
  \vspace{-1mm}
  \caption{
    Illustration of the Next Interleaved Text and Layout Token Prediction objective. 
    The supervision is conducted on both text and layout tokens to reconstruct text content and layout information simultaneously.
      }
  \label{pretrain_figure}
  \vspace{-4mm}
\end{figure}

Leveraging the autoregressive capabilities of LLMs and inspired by the ``Next Token Prediction" in LLMs pre-training, the Next Interleaved Text and Layout Token Prediction (NTLP) is proposed. 
Previous works, such as LayTextLLM~\cite{lu2024boundingboxworthtoken}, focus solely on the prediction of text tokens under interleaved text and layout format without supervising layout information, even though they integrate layout information for LLMs.
Considering the significant role of layout information for document understanding, as illustrated in \cref{pretrain_figure}, NTLP performs the next prediction task to reconstruct the document on all interleaved text and layout tokens, with training on both modalities. 
Thus, NTLP enables effective learning of layout information, enhances cross-modal prediction, and improves relational learning between text and layout modalities.

Specifically, NTLP minimizes the loss between the grounding truth of the next token and its prediction, whether the token is a text token or a layout token, and the loss function is defined as:
\begin{equation}
    \mathcal{L} = \frac{1}{N-1}\sum_{i=1}^{N-1}\mathcal{L}_i(z^i | z^0,z^1,...,z^{i-1}),
  \end{equation}
where \(z^i\) denotes the \(i\)-th token, while \(\mathcal{L}_i\) represents the loss associated with predicting the token \(z^i\) based on all preceding tokens \(z^0, z^1, \ldots, z^{i-1}\).
For supervised training involving text, employing the commonly used cross-entropy (CE) loss associated with large language models (LLMs). 
Notably, given that the layout information has been encoded as a single token alongside the floating-point representation of layout (bounding box) information, NTLP introduces a dedicated layout head \(f_{lay}\) to map the layout hidden states to four-dimensional coordinates \([x_1, y_1, x_2, y_2]\), which serve as the predicted layout output for supervised training utilizing Mean Squared Error (MSE) loss. 
Thus, \(\mathcal{L}_i\) can be expressed as:
\begin{equation} 
    \mathcal{L}_i = 
    \begin{cases}
        \mathcal{L}_{CE}(f_{text}(z^i), y_{text}^i), z^i \in \mathcal{C}_{lay}, \\
        \mathcal{L}_{MSE}(f_{lay}(z^i), Box^i), z^i \in \mathcal{C}_{text},
    \end{cases}
\end{equation}
where \(f_{text}\) denotes the text head, while \(y_{text}^i\) represents the one-hot encoding of the true label corresponding to text token \(z^i \in \mathcal{C}_{text}\). Additionally, \(Box^i\) signifies the true four-dimensional coordinates for layout token \(z^i \in \mathcal{C}_{lay}\).

\section{Experiments}
\subsection{Training Dataset Collection}
\noindent{\bf{Pre-training data}} 
of LayTokenLLM utilizes the open-source document dataset called Layout-aware SFT data from LayoutLLM~\cite{luo2024layoutllm}, which comprises an ensemble of diverse and high-quality data relevant to document understanding and information extraction tasks. 
For pre-training efficiently, filtering out too long documents with token lengths of more than 2k for effective pre-training.

\noindent{\bf{SFT data}} 
of LayTokenLLM employs the datasets extensively used in single-page and multi-page document understanding tasks to ensure high-quality SFT.
For the single-page document understanding task, the combined training sets of DocVQA~\cite{mathew2021docvqa} and SIBR~\cite{yang2023modeling} constitute the SFT dataset. 
DocVQA includes 50k question-answer pairs grounded on 12k document images. Meanwhile, SIBR is a real-world dataset for Visual Information Extraction tasks, covering challenging scenarios with difficult-to-recognize text like blur, partial occlusions, and printing shifts.
Regarding multi-page document understanding SFT, leverages an ensemble of datasets that incorporates the training sets from MP-DocVQA~\cite{tito2023hierarchical} and DUDE~\cite{van2023document}.

\subsection{Training Setup}
In the experiments, two widely used models, Qwen1.5-7B~\cite{qwen1.5} and LLama3-8B~\cite{dubey2024llama}, are employed as the main LLM components of LayTokenLLM, referred to as LayTokenLLM-7B and LayTokenLLM-8B, respectively.
Moreover, for a more comprehensive comparison with other Layout-as-Token methods, we also consider comparisons using the same training data and LLM backbone as ours, but employing different commonly used text and layout formats as input proposed by existing methods~\cite{he2023icl,qwen-vl,internvl}, such as ``\{text:``text",Box:[123, 456, 133, 500]\}", as shown in \cref{table:comparison}.
During both pre-training and SFT phases, as illustrated in \cref{pipeline_figure}, the LLM is frozen, while the parameters of the LoRA~\cite{hu2021lora}, layout tokenizer, and layout head are randomly initialized and updated to support lightweight training.
The pretraining stage and single-page document SFT are trained for 3 epochs with a batch size of 64, a learning rate of 3e-4, and a maximum position ID set to 2048. 
To handle full training of long-context content under computational constraints, multi-page document SFT employs a 2-stage strategy: first, processing documents up to 4k tokens (maximum position ID) with a batch size of 32; second, handling those exceeding 4k up to 16k tokens and batch size is 8.
The training is performed on 8 Nvidia A100 GPUs.

\begin{table*}[htp]
  \small 
  \centering
  \begin{tabular}{
        m{.83\columnwidth}|
        m{.09\columnwidth}<{\centering\arraybackslash}|
        m{.21\columnwidth}<{\centering\arraybackslash}|
        m{.21\columnwidth}<{\centering\arraybackslash}|
        m{.13\columnwidth}<{\centering\arraybackslash}|
        m{.13\columnwidth}<{\centering\arraybackslash}|
        m{.09\columnwidth}<{\centering\arraybackslash}
    }
    \toprule[1pt]
    \multirow{2}*{\textbf{Setting}} & 
    \multicolumn{4}{c|}{\textbf{\makecell{Single-page\\Document VQA}}} & 
    \multicolumn{2}{c}{\textbf{\makecell{Multi-page\\Document VQA}}}  \\
    \cline{2-7}
      & \bf{SIBR} & \bf{FUNSD} & \bf{CORD} & \bf{DocVQA} & \bf{MP-DocVQA} & \bf{DUDE}  \\
    \hline
    \multirow{1}{*}{\makecell{\bf{Plain Text}}} & & & & & & \\
      Qwen1.5-7B-Chat~\cite{qwen1.5}  & 38.81 & 52.52 & 29.71 & 64.27   & 47.15 & 28.98  \\
      Llama3-8B-Instruct~\cite{dubey2024llama}  & 51.77 & 57.47 & 40.00 & 74.22 & 50.75 & 24.89  \\
   \hline
   \multirow{1}{*}{\makecell{\bf{Text + Layout-as-Modality}}} & & & & & &  \\
    DocLLM-7B$\diamond$~\cite{docllm}  & - & (51.80) & (67.40) & 69.50 & - & -  \\
    LayoutLLM-7B$\diamond$~\cite{luo2024layoutllm}  & - & \underline{79.98} & 63.10 & 74.27& - & - \\

  \hline
   \multirow{1}{*}{\makecell{\bf{Text + Layout-as-Token}}} & & & & & \\
        LayTextLLM-7B$\diamond$~\cite{lu2024boundingboxworthtoken} &- & 72.00 & 45.50 &77.20 &- & -  \\
        text, [123, 456, 133, 500]$\star$  &	91.44 &	79.89 &	67.77 &  81.16 & \underline{59.17}	& \underline{41.01}  \\
        \scalebox{0.92}{\{text:``text",Box:[123, 456, 133, 500]\}}$\star$~\cite{he2023icl}  &	\underline{91.45}&	\underline{79.98}	&68.57 & \underline{81.98} &	55.96	&37.96  \\
        \scalebox{0.92}{$<$ref$>$text$<$/ref$><$box$>$(123,456),(133,500)$<$/box$>$}$\star$~\cite{bai2023qwenvl}  &	91.43	&79.56&	\underline{69.62} & 81.37 	&57.81	&39.67 \\
        \scalebox{0.92}{$<$ref$>$text$<$/ref$>$$<$box$>$[[253, 231, 733, 787]]$<$/box$>$}$\star$~\cite{chen2023internvl} 	& 88.24	&78.17	&56.32 & 80.18  &56.16	&40.82\\
        
    \hline

    \hline
    \bf{LayTokenLLM-llama2-7B$\diamond$ (Ours)}  	&90.13	&76.10 (67.39)	&67.60 (73.39) & 79.98 	&56.30	&36.59 \\
    \bf{LayTokenLLM-7B$\star$ (Ours)}  	&92.03	&78.72 (69.47)	&73.79 (71.03)  & 81.50 	&72.81	&49.72 \\
    \bf{LayTokenLLM-8B$\triangle$ (Ours)}  & \bf{92.20} & \bf{81.62 (70.96)} & \bf{78.30 (75.35)} & \bf{85.11}  & \bf{74.31} & \bf{52.00}  \\
    \bottomrule[1pt]
  \end{tabular}
  \vspace{-1mm}
  \caption{
    Comparison with the LLMs integrating layout information. 
    Symbols $\diamond$, $\star$ and $\triangle$ represent the LLM backbones used: Llama2-7B, Qwen1.5-7B and Llama3-8B. Methods marked with $\star$ are trained identically to LayTokenLLM. ($\cdot$) shows F1-scores on uncleaned FUNSD and CORD, as used in DocLLM~\cite{docllm}.
`\textbf{Bold}' means the best in our series, while `\underline{Underline}' marks the best among all compared methods.
  }
  \label{table:main_results}
  \vspace{-2mm}
\end{table*}

\begin{table}[htp]
  \small 
  \centering
  \begin{tabular}{
        m{.32\columnwidth}|
        m{.1\columnwidth}<{\centering\arraybackslash}|
        m{.12\columnwidth}<{\centering\arraybackslash}|
        m{.12\columnwidth}<{\centering\arraybackslash}|
        m{.12\columnwidth}<{\centering\arraybackslash}
    }
    \toprule[1pt]
     \bf{Models} & \bf{SIBR}   & \bf{FUNSD} & \bf{CORD} & \bf{DocVQA} \\
    \hline
        QwenVL-7B~\cite{bai2023qwenvl} & 21.65  & 47.09& 30.00& 65.10\\
        InternVL2-8B~\cite{chen2024far} & \underline{68.39}   & \underline{75.84} & \underline{79.88} & \underline{91.66} \\
        TextMonkey-7B~\cite{liu2024textmonkey} & 51.30  & 65.49 & 67.54 & 66.70 \\
    \hline
    \bf{LayTokenLLM-7B} & 92.03   & 78.72 & 73.79 & 81.50  \\
    \bf{LayTokenLLM-8B} & \bf{92.20}  &\bf{81.62} & \bf{78.30}   & \bf{85.11} \\
    \hline
        
  \end{tabular}
  \caption{
    Comparison with MLLMs on single-page document datasets. `\textbf{Bold}' means the best in our series, while `\underline{Underline}' marks the best among all compared methods.
  }
  \label{table:main_compare_mllms_singlepage}
\end{table}

\begin{table}[htp]
  \small 
  \centering
  \begin{tabular}{
        m{.48\columnwidth}|
        m{.25\columnwidth}<{\centering\arraybackslash}|
        m{.15\columnwidth}<{\centering\arraybackslash}
    }
    \toprule[1pt]
     \bf{Models} & \bf{MP-DocVQA} & \bf{DUDE} \\

  \hline
        LongVA-7B~\cite{zhang2024long} & 60.80 & 38.37 \\
        Idefics3-8B~\cite{laurenccon2024building} & 67.15 & 38.65 \\
        LLaVA-next-interleave-7B~\cite{li2024llava} & 44.87 & 28.03 \\
        InternVL2-8B~\cite{chen2023internvl} & 68.00 & 37.00 \\
        MPLUG-DocOwl2-8B~\cite{hu2024mplug} & \underline{69.42} & \underline{46.77} \\
    \hline
    \bf{LayTokenLLM-7B} & 72.81 & 49.72 \\
    \bf{LayTokenLLM-8B} & \bf{74.31} & \bf{52.00} \\
    \hline
        
  \end{tabular}
  \caption{
    Comparison with MLLMs on multi-page document datasets. 
  }
  \label{table:main_compare_mllms_multipage}
  \vspace{-4mm}
\end{table}

\begin{table*}[h!]
\small
  \centering
  \begin{tabular}{lccc}
  \hline
  \textbf{Text and Layout Format} & \textbf{FLOPs/MACs$\downarrow$} & \textbf{\begin{tabular}[c]{@{}c@{}}\textbf{SP Doc VQA}\\ANLS Avg$\uparrow$\end{tabular}} & \textbf{\begin{tabular}[c]{@{}c@{}}\textbf{MP Doc VQA}\\ANLS Avg$\uparrow$\end{tabular}} \\ \hline
Plain Text (w/o Layout) & 7.95/3.98 & 46.33 & 38.07 \\
\hline
  \scalebox{0.9}{\{text:``text",Box:[123, 456, 133, 500]\}}~\cite{he2023icl}   & 28.76/14.57  & 80.49 & 46.96         \\
  \scalebox{0.9}{$<$ref$>$text$<$/ref$><$box$>$(123,456),(133,500)$<$/box$>$}~\cite{bai2023qwenvl,liu2024textmonkey}    & 28.69/14.34 & 80.50 & 48.74     \\
  \scalebox{0.9}{$<$ref$>$text$<$/ref$>$$<$box$>$[[253, 231, 733, 787]]$<$/box$>$}~\cite{chen2023internvl}  & 32.81/16.40 & 75.73 & 48.49  \\ 
  \hline
  text+layout\_token (\textbf{LayTokenLLM})        & \bf{9.32/5.36}  & \bf{81.51} & \bf{61.27}  \\ \hline
  \end{tabular}
  \vspace{-1mm}
  \caption{Comparison with Layout-as-Token Methods on the multi-page document (MP Doc) understanding tasks, which all initialized from Qwen1.5-7B-Chat. `FLOPs/MACs' denotes the Floating Point Operations Per Second (FLOPs) and the Multiply-Accumulate Operations (MACs) on DocVQA which are broadly used to measure the computational complexity and efficiency. }
  \label{table:main_efficient_result}
  \vspace{-4mm}
\end{table*}

\subsection{Evaluation Setup}
For the single-page document understanding task, widely used benchmarks such as Document Visual Question Answering (Document VQA) and Visual Information Extraction (VIE) are employed, with only the test sets being utilized across all benchmarks. The Document VQA datasets specifically utilize the DocVQA test set, consisting of 5,188 questions. For the VIE task, which includes the SIBR~\cite{yang2023modeling}, FUNSD~\cite{jaume2019funsd}, and CORD~\cite{park2019cord} benchmarks, the cleaned test sets of FUNSD and CORD provided by LayoutLLM~\cite{luo2024layoutllm} are used.
SIBR's test set consists of 400 images, annotated with entity instances and links to challenge visual information extraction models.
The FUNSD dataset features a test collection of 50 form images, each meticulously labeled with entities such as headers, questions, answers, and others, complemented by annotations for entity linking.
Conversely, the CORD dataset encompasses a test suite of 100 receipt images, each enriched with annotations spanning 30 distinct entity categories, including but not limited to tax amounts and total prices.
Following LayoutLLM~\cite{luo2024layoutllm}, transform the VIE datasets into question-answering format, and the QA for both DocVQA and VIE task is evaluated by ANLS~\cite{mathew2021docvqa}.
For the multi-image document understanding task, our experiments test on MP-DocVQA and DUDE, which are widely used for multi-page document understanding.
Following the evaluation metric settings of the original datasets, the MP-DocVQA is evaluated by ANLS, while DUDE adopts a version of ANLS that it has modified.
The hyperparameters during inference (e.g., top k, beam search, etc.) are set to their default values.

\subsection{Main Results}

\subsubsection{Effectiveness Comparison}
\label{sec:effect}
\noindent{\bf{Comparison with LLMs combined with Layout Information}}
is illustrated in \cref{table:main_results}. It can be seen that variant LLMs that incorporate layout information consistently outperform plain text models in all document comprehension tasks, proving that layout is crucial for document understanding.
Moreover, our method achieves competitive results in single-page document VQA (leading in 2 subtasks and with a higher average compared to other methods using the same LLM). Notably, LayTokenLLM outperforms other methods by a large margin in multi-page document VQA (more than 10\% improvement among the marked $\star$ approaches).
We believe the more significant improvement on multi-page document VQA is due to the fact that in single-page documents most cases do not exceed the trained maximum position ID. Consequently, the impact of additional layout information on the LLM can be largely alleviated through further fine-tuning. 
In contrast, in multi-page documents with extensive context that require the allocation of numerous position IDs, the introduction of additional position IDs for layout information may exacerbate the challenges associated with long-context processing. Our LayTokenLLM demonstrates remarkable performance by effectively circumventing the need for extra position IDs dedicated to layout information, thereby emphasizing its efficiency and superiority in handling such complex scenarios.
Furthermore, experiments with different LLM backbone initializations consistently achieve superior results across all benchmarks, substantiating that LayTokenLLM can adapt to various LLMs.

\noindent{\bf{Comparison with MLLMs}}
is shown in \cref{table:main_compare_mllms_singlepage} and \cref{table:main_compare_mllms_multipage}.
Considering the distinct advantages of existing MLLMs in both single-page and multi-page document understanding tasks, representative works in each task are selected for comparison.
It can be seen that LayTokenLLM achieves the comparable performance of the best model InternVL2-8B across most single-page tasks. 
Particularly in challenging scenarios like SIBR, which covers difficult-to-recognize text, LayTokenLLM achieves 92.20\%, compared to InternVL2-8B's 68.39\%, showcasing a significant advantage which is attributed to the enhanced preservation of textual and layout information of the document.
Furthermore, in multi-page document understanding, LayTokenLLM exceeds both InternVL2 and MPLUG-DocOwl by over 5\% on the DUDE dataset. This superiority may stem from MLLMs often compressing images into fewer tokens for multi-page documents, which results in the loss of textual information. In contrast, LayTokenLLM retains a greater proportion of text, enhancing document representation and discernment.

\subsubsection{Efficiency Comparison}
\label{sec:efficieny}
\cref{table:main_efficient_result} presents a comparative analysis of the Layout-as-Token method in terms of efficiency and performance. 
Compared to the methods with a comparable number of parameters, LayTokenLLM demonstrates superior performance in both single-page and multi-page document understanding tasks while exhibiting better efficiency. 
Notably, due to its lightweight design, our LayTokenLLM exhibits a low processing time that is comparable to only Plain Text input, and more than half that of alternative methods integrating layout information. These results affirm that LayTokenLLM is both effective and efficient.

\begin{table}[!t]
  \small
  \begin{center}
      \resizebox{1\linewidth}{!}{
          \begin{tabular}{
              m{.03\columnwidth}<{\centering}|
              m{.14\columnwidth}<{\centering}
              m{.14\columnwidth}<{\centering}
              m{.18\columnwidth}<{\centering}|
              m{.18\columnwidth}<{\centering}|
              m{.18\columnwidth}<{\centering}
          }
          \toprule[1pt]
          \# & \multicolumn{2}{c}{\makecell{Layout Token}} & \multirow{3}[1]{*}{\makecell{NTLP}} & \multirow{3}[1]{*}{\makecell{SP Doc\\ANLS Avg}} & \multirow{3}[1]{*}{\makecell{MP Doc \\ ANLS Avg}}  \\
          \cline{2-3}\noalign{\vskip -2ex}
& \makecell{Layout\\Tokenizer}  & \makecell{LayPosID} & & &  \\\noalign{\vskip -1ex}
          \hline
          0 & & & & 80.07 & 50.09 \\
          1 & \checkmark &  & &  78.89 & 58.22  \\
          2 & \checkmark & \checkmark && 79.27 & 60.50 \\
          3 & \checkmark & \checkmark & \checkmark & \bf{81.51} & \bf{61.27} \\
          \bottomrule[1pt]
          \end{tabular}
      }
  \vspace{-2mm}
  \caption{Ablation study on single-page document (SP Doc) and multi-page document  (MP Doc) understanding tasks.
  LayPosID represents the positional encoding scheme for our Layout Token.}
  \vspace{-5mm}
  \label{table:ablation_results}
  \end{center}
\end{table}

 \begin{figure*}[!tp]
 \centering
  \includegraphics[width=0.95\linewidth]{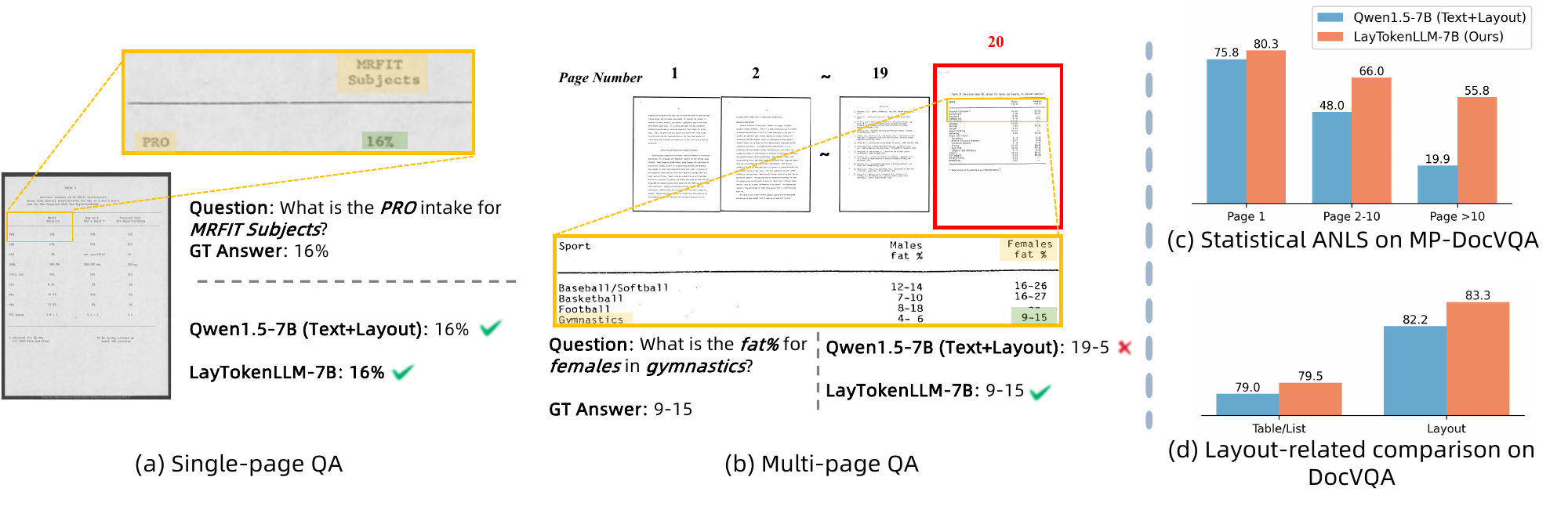}
  \vspace{-4mm}
  \caption{
    Qualitative results on (a) single-page and (b) multi-page document QA, where ``Qwen1.5-7B (Text+Layout)" is trained with the same data and LLM as LayTokenLLM-7B, but employs norm text and layout format (``text, [123, 456, 133, 500]") instead of Layout Token. The \textcolor[RGB]{255,192,0}{\textbf{Yellow}} highlights denote the relevant areas or keys for QA, while the \textcolor[RGB]{77,168,107}{\textbf{Green}} highlights indicate the correct answers. 
    (c) Distribution of statistical ANLS in terms of pages along the posed questions on MP-DocVQA.
    (d) Comparison of layout-related performance using the single-page document dataset, DocVQA.
      }
  \label{fig:case_study}
  \vspace{-4mm}
\end{figure*}
\subsection{Ablation Study}

To evaluate the effectiveness of the proposed Layout Token and pre-training objective in the document understanding task, an ablation study is conducted (see \cref{table:ablation_results}).

\noindent{\bf{Initial Baseline.}} The \#0 baseline disables both \textit{Layout Token} and \textit{NTLP} objective. It utilizes uncompressed layout information as textual tokens for LLM input, consistent with the fine-tuning data as LayTokenLLM settings. Under this configuration, the baseline achieves a high average performance of 80.07\% in single-page document understanding tasks, but performs poorly in multi-page document scenarios due to the layout information occupying critical text learning and understanding space.

\noindent{\bf{Effect of Layout Token.}} 
The proposed Layout Token in LayTokenLLM is generated by our Layout Tokenizer with LayPosID.
In \#1, the layout tokenizer is introduced. Compared to \#0, the proposed compression of layout information enables more text information to be learned within a fixed window, leading to significant performance improvements in multi-page tasks. 
Meanwhile, in single-page document scenarios, there is a slight degradation in performance due to the information loss caused by layout information compression.
In \#2, the framework extends \#1 via LayPosID (our positional encoding scheme), which further eliminates the need for extra layout positional indexing and achieves 100\% T-ratio. 
As a result, \#2 demonstrates additional performance gains over \#1, with a substantial improvement of 2.3\% in multi-page document understanding.

\noindent{\bf{Effect of NTLP Objective.}} Compared with \#2, \#3 further incorporated the \textit{NTLP}, which employs a next interleaved text and layout prediction task. The objective enhances both text and layout representation learning, as well as their interconnections. Performance improvements are observed in both single-page and multi-page document understanding tasks, with increases of 2.2\% and 0.8\% respectively.

Overall, the ablation study confirms the effectiveness of the Layout Token and the NTLP pre-training objective.

\subsection{Qualitative Results}

To further study the effectiveness of our method, two examples from single-page and multi-page document QA scenarios and statistical analysis related to page numbers are presented in \cref{fig:case_study}.
In the context of key-value QAs that rely on spatial layouts, the Qwen1.5-7B model, which integrates standard text and layout formats, can accurately respond on single-page documents (\cref{fig:case_study}(a)) but exhibits answer confusion on multi-page documents (\cref{fig:case_study}(b)). 
In contrast, LayTokenLLM achieves correct reasoning on both single-page and multi-page documents.
We think the confusion in multi-page documents is mainly due to the added position IDs overhead caused by incorporating layout information, leading to long-context issues.
So we further conduct a statistical analysis on the performance related to the page ordinal number with proposed questions, as depicted in \cref{fig:case_study}(c).
It can be seen that the performance of the Qwen1.5-7B model with the direct integration of layout information declines significantly with an increasing number of pages. 
In contrast, our LayTokenLLM exhibits a marked performance advantage as pages increase,  
highlighting its superiority, especially in understanding long-context documents.
Moreover, LayTokenLLM's layout representation performance is further evaluated under conditions excluding the impact of position ID overhead (short-context scenario), using the ``table/list" and ``layout" subset of the DocVQA dataset, see \cref{fig:case_study}(d).  
The results show that LayTokenLLM not only avoids negative impacts but also improves results compared with Qwen1.5-7B (Text+Layout),
demonstrating its effectiveness in re-expressing layout information.
Overall, LayTokenLLM ensures comprehensive text learning while clearly preserving layout information, leading a more complete document understanding.

\section{Limitations}
\label{sec:limitation}
Although the proposed Layout Token demonstrates that LayTokenLLM can effectively address text-dense documents with rich layout information, it may overlook certain graphical elements, such as charts and icons. 
Additionally, although NTLP pre-training has been shown to enhance document understanding, future work could explore more granular tasks, such as fine-grained layout relationship prediction. Further research may focus on equipping LayTokenLLM with these capabilities.

\section{Conclusion}
\label{sec:conclusion}
We propose LayTokenLLM, which incorporates a simple yet effective Layout Token to ensure comprehensive learning of text content while alleviating long-context issues introduced by layout information. 
Furthermore, an interleaved text and layout token next prediction pre-training objective is utilized to enhance cross-modal prediction and relational learning between text and layout modalities.
Extensive experiments demonstrate the effectiveness of LayTokenLLM across diverse benchmarks for 
both single-page and multi-page document understanding.

{
    \small
    \bibliographystyle{ieeenat_fullname}
    \bibliography{main}

\begin{thebibliography}{52}
\providecommand{\natexlab}[1]{#1}
\providecommand{\url}[1]{\texttt{#1}}
\expandafter\ifx\csname urlstyle\endcsname\relax
  \providecommand{\doi}[1]{doi: #1}\else
  \providecommand{\doi}{doi: \begingroup \urlstyle{rm}\Url}\fi

\bibitem[gpt(2023)]{gpt4v}
Gpt-4v(ision) system card.
\newblock 2023.

\bibitem[Appalaraju et~al.(2021)Appalaraju, Jasani, and Kota]{appalaraju2021docformer}
Srikar Appalaraju, Bhavan Jasani, and Bhargava~Urala Kota.
\newblock {DocFormer}: End-to-end transformer for document understanding.
\newblock In \emph{ICCV}, pages 4171--4186, 2021.

\bibitem[Bai et~al.(2023{\natexlab{a}})Bai, Bai, Yang, Wang, Tan, Wang, Lin, Zhou, and Zhou]{bai2023qwenvl}
Jinze Bai, Shuai Bai, Shusheng Yang, Shijie Wang, Sinan Tan, Peng Wang, Junyang Lin, Chang Zhou, and Jingren Zhou.
\newblock Qwen-vl: A frontier large vision-language model with versatile abilities.
\newblock \emph{arXiv preprint arXiv:2308.12966}, 2023{\natexlab{a}}.

\bibitem[Bai et~al.(2023{\natexlab{b}})Bai, Bai, Yang, Wang, Tan, Wang, Lin, Zhou, and Zhou]{qwen-vl}
Jinze Bai, Shuai Bai, Shusheng Yang, Shijie Wang, Sinan Tan, Peng Wang, Junyang Lin, Chang Zhou, and Jingren Zhou.
\newblock Qwen-vl: A versatile vision-language model for understanding, localization, text reading, and beyond.
\newblock 2023{\natexlab{b}}.

\bibitem[Brown et~al.(2020)Brown, Mann, Ryder, Subbiah, Kaplan, Dhariwal, Neelakantan, Shyam, Sastry, Askell, Agarwal, Herbert-Voss, Krueger, Henighan, Child, Ramesh, Ziegler, Wu, Winter, Hesse, Chen, Sigler, Litwin, Gray, Chess, Clark, Berner, McCandlish, Radford, Sutskever, and Amodei]{gpt3}
Tom Brown, Benjamin Mann, Nick Ryder, Melanie Subbiah, Jared~D Kaplan, Prafulla Dhariwal, Arvind Neelakantan, Pranav Shyam, Girish Sastry, Amanda Askell, Sandhini Agarwal, Ariel Herbert-Voss, Gretchen Krueger, Tom Henighan, Rewon Child, Aditya Ramesh, Daniel Ziegler, Jeffrey Wu, Clemens Winter, Chris Hesse, Mark Chen, Eric Sigler, Mateusz Litwin, Scott Gray, Benjamin Chess, Jack Clark, Christopher Berner, Sam McCandlish, Alec Radford, Ilya Sutskever, and Dario Amodei.
\newblock Language models are few-shot learners.
\newblock In \emph{Advances in Neural Information Processing Systems}, pages 1877--1901. Curran Associates, Inc., 2020.

\bibitem[Chen et~al.(2023)Chen, Wu, Wang, Su, Chen, Xing, Zhong, Zhang, Zhu, Lu, Li, Luo, Lu, Qiao, and Dai]{chen2023internvl}
Zhe Chen, Jiannan Wu, Wenhai Wang, Weijie Su, Guo Chen, Sen Xing, Muyan Zhong, Qinglong Zhang, Xizhou Zhu, Lewei Lu, Bin Li, Ping Luo, Tong Lu, Yu Qiao, and Jifeng Dai.
\newblock Internvl: Scaling up vision foundation models and aligning for generic visual-linguistic tasks.
\newblock \emph{arXiv preprint arXiv:2312.14238}, 2023.

\bibitem[Chen et~al.(2024{\natexlab{a}})Chen, Wang, Tian, Ye, Gao, Cui, Tong, Hu, Luo, Ma, et~al.]{chen2024far}
Zhe Chen, Weiyun Wang, Hao Tian, Shenglong Ye, Zhangwei Gao, Erfei Cui, Wenwen Tong, Kongzhi Hu, Jiapeng Luo, Zheng Ma, et~al.
\newblock How far are we to gpt-4v? closing the gap to commercial multimodal models with open-source suites.
\newblock \emph{arXiv preprint arXiv:2404.16821}, 2024{\natexlab{a}}.

\bibitem[Chen et~al.(2024{\natexlab{b}})Chen, Wu, Wang, Su, Chen, Xing, Zhong, Zhang, Zhu, Lu, et~al.]{internvl}
Zhe Chen, Jiannan Wu, Wenhai Wang, Weijie Su, Guo Chen, Sen Xing, Muyan Zhong, Qinglong Zhang, Xizhou Zhu, Lewei Lu, et~al.
\newblock Internvl: Scaling up vision foundation models and aligning for generic visual-linguistic tasks.
\newblock In \emph{Proceedings of the IEEE/CVF Conference on Computer Vision and Pattern Recognition}, pages 24185--24198, 2024{\natexlab{b}}.

\bibitem[Cui et~al.(2021)Cui, Xu, Lv, and Wei]{cui2021document}
Lei Cui, Yiheng Xu, Tengchao Lv, and Furu Wei.
\newblock Document ai: Benchmarks, models and applications.
\newblock \emph{arXiv preprint arXiv:2111.08609}, 2021.

\bibitem[Da et~al.(2023)Da, Luo, Zheng, and Yao]{da2023vision}
Cheng Da, Chuwei Luo, Qi Zheng, and Cong Yao.
\newblock Vision grid transformer for document layout analysis.
\newblock In \emph{ICCV}, pages 19462--19472, 2023.

\bibitem[Dubey et~al.(2024)Dubey, Jauhri, Pandey, Kadian, Al-Dahle, Letman, Mathur, Schelten, Yang, Fan, et~al.]{dubey2024llama}
Abhimanyu Dubey, Abhinav Jauhri, Abhinav Pandey, Abhishek Kadian, Ahmad Al-Dahle, Aiesha Letman, Akhil Mathur, Alan Schelten, Amy Yang, Angela Fan, et~al.
\newblock The llama 3 herd of models.
\newblock \emph{arXiv preprint arXiv:2407.21783}, 2024.

\bibitem[Gu et~al.(2022)Gu, Meng, Wang, Lan, Wang, Gu, and Zhang]{gu2022xylayoutlm}
Zhangxuan Gu, Changhua Meng, Ke Wang, Jun Lan, Weiqiang Wang, Ming Gu, and Liqing Zhang.
\newblock Xylayoutlm: Towards layout-aware multimodal networks for visually-rich document understanding.
\newblock In \emph{CVPR}, pages 4583--4592, 2022.

\bibitem[He et~al.(2023)He, Wang, Hu, Liu, Liu, Xu, and Shen]{he2023icl}
Jiabang He, Lei Wang, Yi Hu, Ning Liu, Hui Liu, Xing Xu, and Heng~Tao Shen.
\newblock Icl-d3ie: In-context learning with diverse demonstrations updating for document information extraction.
\newblock \emph{ICCV}, 2023.

\bibitem[Hu et~al.(2024{\natexlab{a}})Hu, Xu, Ye, Yan, Zhang, Zhang, Li, Zhang, Jin, Huang, et~al.]{hu2024mplug1.5}
Anwen Hu, Haiyang Xu, Jiabo Ye, Ming Yan, Liang Zhang, Bo Zhang, Chen Li, Ji Zhang, Qin Jin, Fei Huang, et~al.
\newblock mplug-docowl 1.5: Unified structure learning for ocr-free document understanding.
\newblock \emph{arXiv preprint arXiv:2403.12895}, 2024{\natexlab{a}}.

\bibitem[Hu et~al.(2024{\natexlab{b}})Hu, Xu, Zhang, Ye, Yan, Zhang, Jin, Huang, and Zhou]{hu2024mplug}
Anwen Hu, Haiyang Xu, Liang Zhang, Jiabo Ye, Ming Yan, Ji Zhang, Qin Jin, Fei Huang, and Jingren Zhou.
\newblock mplug-docowl2: High-resolution compressing for ocr-free multi-page document understanding.
\newblock \emph{arXiv preprint arXiv:2409.03420}, 2024{\natexlab{b}}.

\bibitem[Hu et~al.(2021)Hu, Shen, Wallis, Allen-Zhu, Li, Wang, Wang, and Chen]{hu2021lora}
Edward~J Hu, Yelong Shen, Phillip Wallis, Zeyuan Allen-Zhu, Yuanzhi Li, Shean Wang, Lu Wang, and Weizhu Chen.
\newblock Lora: Low-rank adaptation of large language models.
\newblock \emph{arXiv preprint arXiv:2106.09685}, 2021.

\bibitem[Huang et~al.(2022{\natexlab{a}})Huang, Lv, Cui, Lu, and Wei]{huang2022layoutlmv3}
Yupan Huang, Tengchao Lv, Lei Cui, Yutong Lu, and Furu Wei.
\newblock Layoutlmv3: Pre-training for document ai with unified text and image masking.
\newblock In \emph{ACM Multimedia}, 2022{\natexlab{a}}.

\bibitem[Huang et~al.(2022{\natexlab{b}})Huang, Lv, Cui, Lu, and Wei]{layoutlmv3}
Yupan Huang, Tengchao Lv, Lei Cui, Yutong Lu, and Furu Wei.
\newblock Layoutlmv3: Pre-training for document ai with unified text and image masking.
\newblock In \emph{Proceedings of the 30th ACM International Conference on Multimedia}, pages 4083--4091, New York, NY, USA, 2022{\natexlab{b}}. Association for Computing Machinery.

\bibitem[Jaume et~al.(2019)Jaume, Ekenel, and Thiran]{jaume2019funsd}
Guillaume Jaume, Hazim~Kemal Ekenel, and Jean-Philippe Thiran.
\newblock Funsd: A dataset for form understanding in noisy scanned documents, 2019.

\bibitem[Lamott et~al.(2024)Lamott, Weweler, Ulges, Shafait, Krechel, and Obradovic]{lapdoc}
Marcel Lamott, Yves-Noel Weweler, Adrian Ulges, Faisal Shafait, Dirk Krechel, and Darko Obradovic.
\newblock Lapdoc: Layout-aware prompting for documents.
\newblock 2024.

\bibitem[Lauren{\c{c}}on et~al.(2024)Lauren{\c{c}}on, Marafioti, Sanh, and Tronchon]{laurenccon2024building}
Hugo Lauren{\c{c}}on, Andr{\'e}s Marafioti, Victor Sanh, and L{\'e}o Tronchon.
\newblock Building and better understanding vision-language models: insights and future directions.
\newblock \emph{arXiv preprint arXiv:2408.12637}, 2024.

\bibitem[Li et~al.(2021{\natexlab{a}})Li, Bi, and Yan]{li2021structurallm}
Chenliang Li, Bin Bi, and Ming Yan.
\newblock {StructuralLM}: Structural pre-training for form understanding.
\newblock In \emph{ACL}, 2021{\natexlab{a}}.

\bibitem[Li et~al.(2024{\natexlab{a}})Li, Zhang, Zhang, Zhang, Li, Li, Ma, and Li]{li2024llava}
Feng Li, Renrui Zhang, Hao Zhang, Yuanhan Zhang, Bo Li, Wei Li, Zejun Ma, and Chunyuan Li.
\newblock Llava-next-interleave: Tackling multi-image, video, and 3d in large multimodal models.
\newblock \emph{arXiv preprint arXiv:2407.07895}, 2024{\natexlab{a}}.

\bibitem[Li et~al.(2024{\natexlab{b}})Li, Yuan, Liu, Tang, Wang, Zhu, and Zhang]{li2024tokenpacker}
Wentong Li, Yuqian Yuan, Jian Liu, Dongqi Tang, Song Wang, Jianke Zhu, and Lei Zhang.
\newblock Tokenpacker: Efficient visual projector for multimodal llm.
\newblock \emph{arXiv preprint arXiv:2407.02392}, 2024{\natexlab{b}}.

\bibitem[Li et~al.(2021{\natexlab{b}})Li, Qian, Yu, Qin, Zhang, Liu, Yao, Han, Liu, and Ding]{li2021structext}
Yulin Li, Yuxi Qian, Yuechen Yu, Xiameng Qin, Chengquan Zhang, Yan Liu, Kun Yao, Junyu Han, Jingtuo Liu, and Errui Ding.
\newblock Structext: Structured text understanding with multi-modal transformers.
\newblock In \emph{ACM Multimedia}, pages 1912--1920, 2021{\natexlab{b}}.

\bibitem[Liu et~al.(2023)Liu, Li, Wu, and Lee]{NEURIPS2023_6dcf277e}
Haotian Liu, Chunyuan Li, Qingyang Wu, and Yong~Jae Lee.
\newblock Visual instruction tuning.
\newblock In \emph{Advances in Neural Information Processing Systems}, pages 34892--34916. Curran Associates, Inc., 2023.

\bibitem[Liu et~al.(2024)Liu, Yang, Liu, Li, Ma, Zhang, and Bai]{liu2024textmonkey}
Yuliang Liu, Biao Yang, Qiang Liu, Zhang Li, Zhiyin Ma, Shuo Zhang, and Xiang Bai.
\newblock Textmonkey: An ocr-free large multimodal model for understanding document.
\newblock \emph{arXiv preprint arXiv:2403.04473}, 2024.

\bibitem[Lu et~al.(2024)Lu, Yu, Wang, Ye, Tang, Yang, Wu, Liu, Feng, Wang, Liu, and Huang]{lu2024boundingboxworthtoken}
Jinghui Lu, Haiyang Yu, Yanjie Wang, Yongjie Ye, Jingqun Tang, Ziwei Yang, Binghong Wu, Qi Liu, Hao Feng, Han Wang, Hao Liu, and Can Huang.
\newblock A bounding box is worth one token: Interleaving layout and text in a large language model for document understanding, 2024.

\bibitem[Luo et~al.(2022)Luo, Tang, Zheng, Yao, Jin, Li, Xue, and Si]{bivldoc}
Chuwei Luo, Guozhi Tang, Qi Zheng, Cong Yao, Lianwen Jin, Chenliang Li, Yang Xue, and Luo Si.
\newblock Bi-vldoc: Bidirectional vision-language modeling for visually-rich document understanding.
\newblock \emph{arXiv preprint arXiv:2206.13155}, 2022.

\bibitem[Luo et~al.(2023)Luo, Cheng, Zheng, and Yao]{Luo_2023_CVPR}
Chuwei Luo, Changxu Cheng, Qi Zheng, and Cong Yao.
\newblock Geolayoutlm: Geometric pre-training for visual information extraction.
\newblock In \emph{CVPR}, pages 7092--7101, 2023.

\bibitem[Luo et~al.(2024)Luo, Shen, Zhu, Zheng, Yu, and Yao]{luo2024layoutllm}
Chuwei Luo, Yufan Shen, Zhaoqing Zhu, Qi Zheng, Zhi Yu, and Cong Yao.
\newblock Layoutllm: Layout instruction tuning with large language models for document understanding.
\newblock In \emph{Proceedings of the IEEE/CVF Conference on Computer Vision and Pattern Recognition}, pages 15630--15640, 2024.

\bibitem[Mathew et~al.(2021)Mathew, Karatzas, and Jawahar]{mathew2021docvqa}
Minesh Mathew, Dimosthenis Karatzas, and CV Jawahar.
\newblock Docvqa: A dataset for vqa on document images.
\newblock In \emph{WACV}, pages 2200--2209, 2021.

\bibitem[Park et~al.(2019)Park, Shin, Lee, Lee, Surh, Seo, and Lee]{park2019cord}
Seunghyun Park, Seung Shin, Bado Lee, Junyeop Lee, Jaeheung Surh, Minjoon Seo, and Hwalsuk Lee.
\newblock {\{}CORD{\}}: A consolidated receipt dataset for post-{\{}ocr{\}} parsing.
\newblock In \emph{Workshop on Document Intelligence at NeurIPS 2019}, 2019.

\bibitem[Peng et~al.(2023)Peng, Quesnelle, Fan, and Shippole]{peng2023yarn}
Bowen Peng, Jeffrey Quesnelle, Honglu Fan, and Enrico Shippole.
\newblock Yarn: Efficient context window extension of large language models.
\newblock \emph{arXiv preprint arXiv:2309.00071}, 2023.

\bibitem[Perot et~al.(2024)Perot, Kang, Luisier, Su, Sun, Boppana, Wang, Wang, Mu, Zhang, Lee, and Hua]{perot2024lmdx}
Vincent Perot, Kai Kang, Florian Luisier, Guolong Su, Xiaoyu Sun, Ramya~Sree Boppana, Zilong Wang, Zifeng Wang, Jiaqi Mu, Hao Zhang, Chen-Yu Lee, and Nan Hua.
\newblock Lmdx: Language model-based document information extraction and localization, 2024.

\bibitem[Shen et~al.(2024)Shen, Luo, Zhu, Chen, Zheng, Yu, Bu, and Yao]{proctag}
Yufan Shen, Chuwei Luo, Zhaoqing Zhu, Yang Chen, Qi Zheng, Zhi Yu, Jiajun Bu, and Cong Yao.
\newblock Proctag: Process tagging for assessing the efficacy of document instruction data.
\newblock \emph{arXiv preprint arXiv:2407.12358}, 2024.

\bibitem[Song et~al.(2024)Song, Oh, Mo, Kim, Yun, Ha, and Shin]{song2024hierarchical}
Woomin Song, Seunghyuk Oh, Sangwoo Mo, Jaehyung Kim, Sukmin Yun, Jung-Woo Ha, and Jinwoo Shin.
\newblock Hierarchical context merging: Better long context understanding for pre-trained llms.
\newblock \emph{arXiv preprint arXiv:2404.10308}, 2024.

\bibitem[Su et~al.(2024)Su, Ahmed, Lu, Pan, Bo, and Liu]{su2024roformer}
Jianlin Su, Murtadha Ahmed, Yu Lu, Shengfeng Pan, Wen Bo, and Yunfeng Liu.
\newblock Roformer: Enhanced transformer with rotary position embedding.
\newblock \emph{Neurocomputing}, 568:\penalty0 127063, 2024.

\bibitem[Team et~al.(2023)Team, Anil, Borgeaud, Wu, Alayrac, Yu, Soricut, Schalkwyk, Dai, Hauth, et~al.]{team2023gemini}
Gemini Team, Rohan Anil, Sebastian Borgeaud, Yonghui Wu, Jean-Baptiste Alayrac, Jiahui Yu, Radu Soricut, Johan Schalkwyk, Andrew~M Dai, Anja Hauth, et~al.
\newblock Gemini: a family of highly capable multimodal models.
\newblock \emph{arXiv preprint arXiv:2312.11805}, 2023.

\bibitem[Team(2024)]{qwen1.5}
Qwen Team.
\newblock Introducing qwen1.5, 2024.

\bibitem[Tito et~al.(2023)Tito, Karatzas, and Valveny]{tito2023hierarchical}
Rub{\`e}n Tito, Dimosthenis Karatzas, and Ernest Valveny.
\newblock Hierarchical multimodal transformers for multipage docvqa.
\newblock \emph{Pattern Recognition}, 144:\penalty0 109834, 2023.

\bibitem[Touvron et~al.(2023)Touvron, Lavril, Izacard, Martinet, Lachaux, Lacroix, Rozi{\`e}re, Goyal, Hambro, Azhar, et~al.]{llama}
Hugo Touvron, Thibaut Lavril, Gautier Izacard, Xavier Martinet, Marie-Anne Lachaux, Timoth{\'e}e Lacroix, Baptiste Rozi{\`e}re, Naman Goyal, Eric Hambro, Faisal Azhar, et~al.
\newblock Llama: Open and efficient foundation language models.
\newblock \emph{arXiv preprint arXiv:2302.13971}, 2023.

\bibitem[Van~Landeghem et~al.(2023)Van~Landeghem, Tito, Borchmann, Pietruszka, Joziak, Powalski, Jurkiewicz, Coustaty, Anckaert, Valveny, et~al.]{van2023document}
Jordy Van~Landeghem, Rub{\`e}n Tito, {\L}ukasz Borchmann, Micha{\l} Pietruszka, Pawel Joziak, Rafal Powalski, Dawid Jurkiewicz, Micka{\"e}l Coustaty, Bertrand Anckaert, Ernest Valveny, et~al.
\newblock Document understanding dataset and evaluation (dude).
\newblock In \emph{Proceedings of the IEEE/CVF International Conference on Computer Vision}, pages 19528--19540, 2023.

\bibitem[Wang et~al.(2024)Wang, Raman, Sibue, Ma, Babkin, Kaur, Pei, Nourbakhsh, and Liu]{docllm}
Dongsheng Wang, Natraj Raman, Mathieu Sibue, Zhiqiang Ma, Petr Babkin, Simerjot Kaur, Yulong Pei, Armineh Nourbakhsh, and Xiaomo Liu.
\newblock {D}oc{LLM}: A layout-aware generative language model for multimodal document understanding.
\newblock In \emph{Proceedings of the 62nd Annual Meeting of the Association for Computational Linguistics (Volume 1: Long Papers)}, pages 8529--8548, Bangkok, Thailand, 2024. Association for Computational Linguistics.

\bibitem[Xu et~al.(2020)Xu, Li, Cui, and Huang]{xu2020layoutlm}
Yiheng Xu, Minghao Li, Lei Cui, and Shaohan Huang.
\newblock {LayoutLM}: Pre-training of text and layout for document image understanding.
\newblock In \emph{KDD}, pages 1192--1200, 2020.

\bibitem[Xu et~al.(2021{\natexlab{a}})Xu, Xu, Lv, Cui, Wei, Wang, Lu, Florencio, Zhang, Che, Zhang, and Zhou]{Xu2020LayoutLMv2MP}
Yang Xu, Yiheng Xu, Tengchao Lv, Lei Cui, Furu Wei, Guoxin Wang, Yijuan Lu, Dinei Florencio, Cha Zhang, Wanxiang Che, Min Zhang, and Lidong Zhou.
\newblock Layoutlmv2: Multi-modal pre-training for visually-rich document understanding.
\newblock In \emph{Proceedings of the 59th Annual Meeting of the Association for Computational Linguistics (ACL) 2021}, 2021{\natexlab{a}}.

\bibitem[Xu et~al.(2021{\natexlab{b}})Xu, Xu, Lv, Cui, Wei, Wang, Lu, Florencio, Zhang, Che, Zhang, and Zhou]{xu2020layoutlmv2}
Yang Xu, Yiheng Xu, Tengchao Lv, Lei Cui, Furu Wei, Guoxin Wang, Yijuan Lu, Dinei Florencio, Cha Zhang, Wanxiang Che, Min Zhang, and Lidong Zhou.
\newblock Layoutlmv2: Multi-modal pre-training for visually-rich document understanding.
\newblock In \emph{ACL}, 2021{\natexlab{b}}.

\bibitem[Yang et~al.(2023)Yang, Long, Wang, Song, Zhong, Cheng, Bai, and Yao]{yang2023modeling}
Zhibo Yang, Rujiao Long, Pengfei Wang, Sibo Song, Humen Zhong, Wenqing Cheng, Xiang Bai, and Cong Yao.
\newblock Modeling entities as semantic points for visual information extraction in the wild.
\newblock In \emph{Proceedings of the IEEE/CVF Conference on Computer Vision and Pattern Recognition}, 2023.

\bibitem[Ye et~al.(2024)Ye, Xu, Ye, Yan, Hu, Liu, Qian, Zhang, and Huang]{ye2024mplug}
Qinghao Ye, Haiyang Xu, Jiabo Ye, Ming Yan, Anwen Hu, Haowei Liu, Qi Qian, Ji Zhang, and Fei Huang.
\newblock mplug-owl2: Revolutionizing multi-modal large language model with modality collaboration.
\newblock In \emph{Proceedings of the IEEE/CVF Conference on Computer Vision and Pattern Recognition}, pages 13040--13051, 2024.

\bibitem[Yu et~al.(2023)Yu, Li, Zhang, Zhang, Guo, Qin, Yao, Han, Ding, and Wang]{yu2023structextv}
Yuechen Yu, Yulin Li, Chengquan Zhang, Xiaoqiang Zhang, Zengyuan Guo, Xiameng Qin, Kun Yao, Junyu Han, Errui Ding, and Jingdong Wang.
\newblock Structextv2: Masked visual-textual prediction for document image pre-training.
\newblock In \emph{ICLR}, 2023.

\bibitem[Zhang et~al.(2024)Zhang, Zhang, Li, Zeng, Yang, Zhang, Wang, Tan, Li, and Liu]{zhang2024long}
Peiyuan Zhang, Kaichen Zhang, Bo Li, Guangtao Zeng, Jingkang Yang, Yuanhan Zhang, Ziyue Wang, Haoran Tan, Chunyuan Li, and Ziwei Liu.
\newblock Long context transfer from language to vision.
\newblock \emph{arXiv preprint arXiv:2406.16852}, 2024.

\bibitem[Zhu et~al.(2024)Zhu, Chen, Shen, Li, and Elhoseiny]{zhu2024minigpt}
Deyao Zhu, Jun Chen, Xiaoqian Shen, Xiang Li, and Mohamed Elhoseiny.
\newblock Mini{GPT}-4: Enhancing vision-language understanding with advanced large language models.
\newblock In \emph{The Twelfth International Conference on Learning Representations}, 2024.

\end{thebibliography}
}


\end{document}